\documentclass[sigconf]{aamas}  
\usepackage{balance}  

\usepackage{booktabs}
\usepackage{todonotes}
\usepackage{cleveref}
\usepackage{xcolor}
\usepackage{float}
\usepackage{subcaption}

\setcopyright{ifaamas}  
\acmDOI{}  
\acmISBN{}  
\acmConference[AAMAS'19]{Proc.\@ of the 18th International Conference on Autonomous Agents and Multiagent Systems (AAMAS 2019)}{May 13--17, 2019}{Montreal, Canada}{N.~Agmon, M.~E.~Taylor, E.~Elkind, M.~Veloso (eds.)}  
\acmYear{2019}  
\copyrightyear{2019}  
\acmPrice{}  

\begin{document}

\title{Object Exchangeability in Reinforcement Learning}  

\subtitle{Extended Abstract}

\author{John Mern}
\affiliation{%
  \institution{Stanford University}
  \city{Stanford} 
  \state{California} 
  \postcode{94305}
}
\email{jmern91@stanford.edu}
\author{Dorsa Sadigh}
\affiliation{%
  \institution{Stanford University}
  \city{Stanford} 
  \state{California} 
  \postcode{94305}
}
\email{dorsa@stanford.edu}
\author{Mykel Kochenderfer}
\affiliation{%
  \institution{Stanford University}
  \city{Stanford} 
  \state{California} 
  \postcode{94305}
}
\email{mykel@stanford.edu}

\begin{abstract}  
Although deep reinforcement learning has advanced significantly over the past several years, sample efficiency remains a major challenge. 
Careful choice of input representations can help improve efficiency depending on the structure present in the problem. In this work, we present an \emph{attention-based} method to project inputs into an efficient representation space that is invariant under changes to input ordering. 
We show that our proposed representation results in a search space that is a factor of $m!$ smaller for inputs of $m$ objects. 
Our experiments demonstrate improvements in sample efficiency for policy gradient methods on a variety of tasks.
We show that our representation allows us to solve problems that are otherwise intractable when using na\"ive approaches.
\end{abstract}

%

\keywords{Knowledge Representation; Reasoning; Reinforcement Learning}  

\maketitle


\section{Introduction}
Deep reinforcement learning (RL) has achieved state-of-the-art performance across a variety of tasks~\cite{Mnih2013, Silver2017}. 
However, successful deep RL training requires large amounts of sample data. 
Various learning methods have been proposed to improve sample efficiency, such as model-based learning and incorporation of Bayesian priors~\cite{Gu2016, Spector2018}.

The key insight of this paper is that we can significantly improve efficiency by leveraging the exchangeable structure inherent in many reinforcement learning problems. 
That is, for a state space that can be factored into sets of sub-states, presenting the factored state in a way that does not rely on a particular ordering of the sub-states can lead to significant reduction in the search-space. 

In this work, we propose an attention mechanism as a means to leverage object exchangeability. 
We propose a mechanism that is permutation invariant in that it will produce the same output for any permutation of the items in the input set and show that this representation reduces the input search space by a factor of up to $m!$, where $m$ is the number of exchangeable objects. 


\section{Background and Related Work}

Deep RL is a class of methods to solve Markov Decision Processes (MDPs) using deep neural networks. 
Solving an MPD requires finding a policy $\pi$ that maps all states in a state-space $\mathcal{S}$ to an action to maximize total accumulated rewards.

Formally, we can define an \emph{object} in an MDP to be a subset of the state space that defines the state of a single entity in the problem environment. In an aircraft collision avoidance problem, an object could be defined by the values associated with a single aircraft.
It is well known that as the number of objects grow, the size of the MDP search space grows exponentially~\cite{Robbel2016}. 

For MDPs with exchangeable objects, an optimal policy should provide the same action for any permutation of the input. 
When states are represented as ordered sets, as is common, this must be learned by the policy during training. 

Many methods have been proposed instead to enforce this by permutation invariant input representations. 
The Object Oriented MDP (OO-MDP) framework uses object-class exchangeability to represent states in an order-invariant space for discrete spaces~\cite{Diuk2008}. 
Approximately Optimal State Abstractions~\cite{Abel2016} proposes a theoretical approximation to extend OO-MDP to continuous domains.  
Object-Focused Q-learning~\cite{Cobo2013} uses object classes to decompose the Q-function output space, though it does not address the input. 

Deep Sets~\cite{Zaheer2017} proposes a permutation invariant abstraction method to produce input vectors from exchangeable sets. 
The method proposed produces a static mapping. That is each input object is weighted equally regardless of value during the mapping. 

 Our method improves upon Deep Sets by introducing an attention mechanism to dynamically map the inputs to the permutation-invariant space. 
 Attention mechanisms are used in various deep learning tasks to dynamically filter the input to a down-stream neural network to emphasize most important parts of the original input~\cite{xu2015, luong2015, jaderberg2015}. 
We adapt a mechanism from recent work in natural language processing, which use a dot-product neural layer to efficiently apply dynamic attention~\cite{Vaswani2017}. 

\section{Problem Formalization}\label{sec:Method}
\begin{figure}[!h]
\centering
\includegraphics[width=.9\columnwidth]{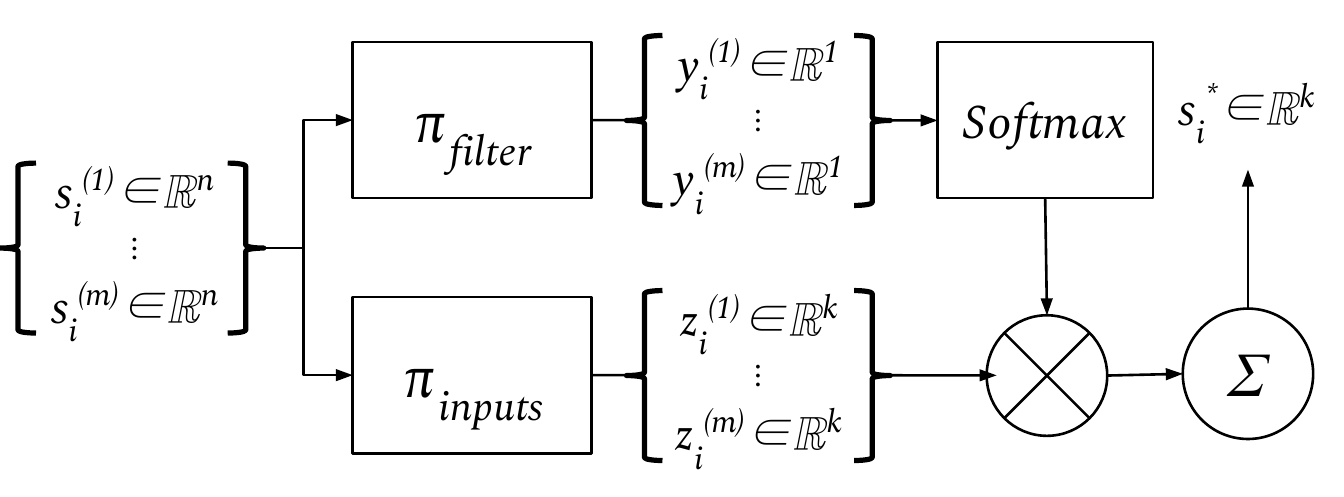}
\caption{Permutation invariant attention mechanism. - Objects $\{s_i^{(1)}, \ldots, s_i^{(m)}\}$ are passed into neural networks $\pi_\text{inputs}$ and $\pi_\text{filters}$ with a Softmax. The outputs are multiplied and summed along the object dimension for the final order-invariant output $s_i^*$. 
} 
\label{fig:attn}
\end{figure}
\begin{figure}
\includegraphics[width=.9\columnwidth]{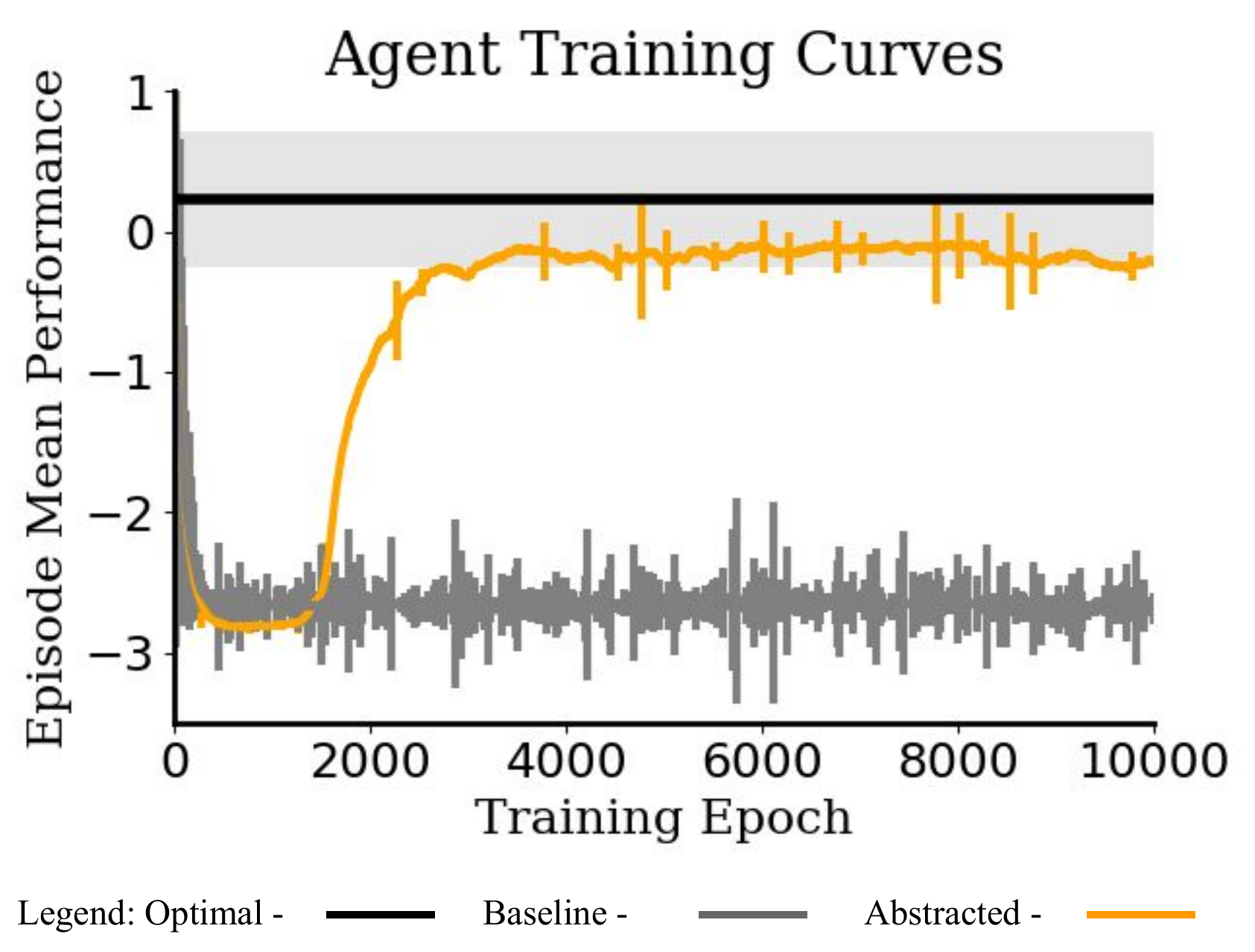}
\caption{Convoy Task Training Curve
} 
\label{fig:convoy}
\end{figure}
Our objective is to propose an attention mechanism that will take sets of objects $S_i$ as inputs and produce abstractions $S_i^*$ such that the mapping is permutation invariant. 
This output can then be used as an input to the policy neural network in an RL problem (e.g. a deep Q-Network or action policy). 
Our hypothesis is that learning using this abstract representation will be more sample efficient than learning on the original object set. 

We propose the attention network architecture shown in~\cref{fig:attn}, which is a permutation invariant implementation of dot-product attention. 
For a single input set $S_i \leftarrow \{s_i^{(1)}, \ldots, s_i^{(m)}\}$, the $m$ object state vectors are individually passed through feed-forward neural networks $\pi_{\text{filter}}$ and $\pi_{\text{inputs}}$. 
The scalar outputs of the filter graph are concatenated into a single vector $y_i \in \mathbb{R}^m$ and the \emph{softmax} operation is applied. 
These outputs are then multiplied element-wise by the concatenated outputs of the network $\pi_{\text{inputs}}$. 
In this way, the output of $\pi_{\text{filter}}$ acts as the attention filter, weighting the inputs by importance prior to summation. 
The elements of the weighted vector $z_i$ are then summed over the $m$ different objects, resulting in a single vector $s^*_i \in \mathbb{R}^k$.
This $s^*_i$ vector is then used as the input to the policy neural network.


We can now define bounds on the sample efficiency benefits of an invariant mapping. Define a state space $\mathcal{S}$ such that $\{s_1, \ldots, s_m\} \in \mathcal{S}$, where $m$ is the number of objects. Let each object $s_i$ take on $n$ unique values. Representing the states as \emph{ordered} sets of $s_i$ results in a state-space size $|\mathcal{S}|$ that can be calculated from the expression for $m$ permutations of $n$ values. If all objects are exchangeable, there exists an abstraction that is permutation invariant. Since the order does not matter, the size of this abstract state $|\hat{\mathcal{S}}|$ can then be calculated from the expression for $m$ combinations of $n$ values. 
\begin{equation}
|\mathcal{S}| = \frac{n!}{(n-m)!}, \ |\hat{\mathcal{S}}| = \frac{n!}{m!(n-m)!}
\end{equation}
Using permutation invariant representation reduces the input space by a factor of $\frac{|S|}{|\hat{S}|} = \frac{1}{m!}$ compared to ordered set representation. 

It can be shown that it is necessary and sufficient for a mapping $f$ to be invariant on all countable sets $\mathcal{X}$ if and only if it can be decomposed using transformations $\phi$ and $\rho$, where $\phi$ and $\rho$ are any vector valued functions 
to the form~\cite{Zaheer2017}:
\begin{equation}\label{eq:invariant}
    f(X) = \rho\Big( \sum_{x\in\mathcal{X}} \phi(x)\Big)
\end{equation}
It can be shown that the proposed attention mechanism may be decomposed to the above form to prove it is permutation invariant. For problems with multiple classes of exchangeable objects, a separate attention mechanism can be deployed for each class. 

\section{Experiments and Results}
We conducted a series of experiments to validate the effectiveness of our proposed abstraction. 
The first two tasks are simple MDPs in which a scavenger agent navigates a continuous two-dimensional world to find \emph{food} particles. 

In the scavenger tasks, the state space contains vectors $s \in \mathbb{R}^{2m+2}$, where $m$ is the number of target objects. 
The vector contains the relative position of each food particle as well as the ego position of the agent. 
The agent receives a reward of $+1.0$ when reaching a food particle, and a reward of $-0.05$ for every time-step otherwise. 
The episode terminates upon reaching a food particle or when the number of time-steps exceeds a limit. 
Scavenger Task 2 introduced \emph{poison} particles in addition to the food particles (one poison for each food particle). 
If an agent reaches a poison particle, a reward of $-1.0$ is given and the episode terminates.

The third task is a convoy protection task with variable numbers of objects. The task requires a defender agent to protect a convoy that follows a predetermined path through a 2D environment. Attackers are spawned at the periphery of the environment during the episode, and the defender must block their attempts to approach the convoy. 
The state space is the space of vectors representing the state of each non-ego object in the environment.
The episode terminates when all convoy members either reach the goal position or are reached by an attacker. 

For each task, we trained a set of policies with the attention mechanism as well as a baseline policies that use a standard ordered set to represent the input space. 
Each policy was trained with Proximal Policy Optimization (PPO)~\cite{Schulman2017}, policy-gradient algorithm.

For each scavenger task, we trained a policy for on tasks having one to five food particles. 
The baseline policies were unable to achieve optimal performance for tasks with more than two food particles in either scavenger task. 
The policy trained with our attention mechanism was able to learn an optimal policy for all cases with no increase in the number of required training samples. For the convoy task, the abstracted policy approached optimal behavior after approximately 2,500 epochs, where the baseline policy showed no improvement after 10,000 epochs, as shown in~\cref{fig:convoy}.

These experiments demonstrate the effectiveness of the proposed approach to enhance the scalability of the PPO policy gradient learning algorithm.
Together, these experiments validate our hypothesis that leveraging object exchangability for input representation can improve the efficiency of deep reinforcement learning.

\bibliographystyle{ACM-Reference-Format}  
\balance  
\bibliography{main}  

\end{document}